\documentclass{article}

\usepackage{microtype}
\usepackage{graphicx}
\usepackage{subfigure}
\usepackage{booktabs} %

\usepackage{hyperref}

\usepackage[accepted]{icml2024}

\usepackage{amsmath}
\usepackage{amssymb}
\usepackage{mathtools}
\usepackage{amsthm}

\usepackage[utf8]{inputenc} %
\usepackage[T1]{fontenc}    %
\usepackage{url}            %
\usepackage{booktabs}       %
\usepackage{amsfonts}       %
\usepackage{nicefrac}       %
\usepackage{microtype}      %
\usepackage{xcolor}         %
\usepackage{amsmath}
\usepackage{amsthm}
\usepackage{cleveref}

\usepackage{soul}
\usepackage{graphicx}
\newcommand{\D}{\mathcal{D}}

\renewcommand{\div}{\operatorname{div}}

\theoremstyle{definition}
\newtheorem{prop}{Proposition}[section]

\theoremstyle{definition}

\theoremstyle{remark}

\usepackage[textsize=tiny]{todonotes}

\icmltitlerunning{Training Data Protection with Compositional Diffusion Models}

\begin{document}

\twocolumn[
\icmltitle{Training Data Protection with Compositional Diffusion Models}

\icmlsetsymbol{equal}{*}

\begin{icmlauthorlist}
\icmlauthor{Aditya Golatkar}{AWS AI Labs}
\icmlauthor{Alessandro Achille}{AWS AI Labs}
\icmlauthor{Ashwin Swaminathan}{AWS AI Labs}
\icmlauthor{Stefano Soatto}{AWS AI Labs}
\end{icmlauthorlist}

\icmlaffiliation{AWS AI Labs}{AWS AI Labs}

\icmlcorrespondingauthor{Aditya Golatkar}{agolatka@amazon.com}

\icmlkeywords{Machine Learning, ICML}

\vskip 0.3in
]

\printAffiliationsAndNotice{}  %

\begin{abstract}
We introduce Compartmentalized Diffusion Models (CDM), a method to train different diffusion models (or prompts) on distinct data sources and arbitrarily compose them at inference time. The individual models can be trained in isolation, at different times, and on different distributions and domains and can be later composed to achieve performance comparable to a paragon model trained on all data simultaneously. Furthermore, each model only contains information about the subset of the data it was exposed to during training, enabling several forms of training data protection. In particular, CDMs enable \textit{perfect} selective forgetting 
and continual learning for large-scale diffusion models, allow serving customized models based on the user's access rights. Empirically the quality (FID) of the class-conditional CDMs (8-splits) is within 10\% (on fine-grained vision datasets) of a monolithic model (no splits), and allows (8x) faster forgetting compared monolithic model with a maximum FID increase of 1\%. When applied to text-to-image generation, CDMs improve alignment (TIFA) by 14.33\% over a monolithic model trained on MSCOCO. CDMs also allow determining the importance of a subset of the data (attribution) in generating particular samples, and reduce memorization.
\end{abstract}

\section{Introduction}

Diffusion models have captured the popular imagination by enabling users to generate compelling images using simple text prompts or sketches. They have also, in some cases, captured the personal workmanship of artists, since the sheer volume of training data makes it challenging to verify each sample's attribution \cite{vyas2023provable}. It is also challenging to quantify the data contribution in shaping the model's generated output, which calls for the development of new forms of protection for large-scale training data, ranging from methods that limit  the influence of training samples \textit{a-priori} (e.g., \textit{differential privacy}), remove the influence of training examples that were wrongly included in the training \textit{a-posteriori} (\textit{selective forgetting}, \textit{model disgorgement}), and limit the influence of samples on the training output (\textit{copyright protection}), or at least identify which samples had the most influence (\textit{attribution}), thus preventing memorization and/or generation of samples that are substantially similar to training data.
While research in these fields is thriving, the methods developed are not transferable to large-scale diffusion models. Extending known techniques seems daunting since information from different samples is mixed irreversibly the weights of the model, making unlearning or evaluating the influence of specific data challenging. 

We introduce Compartmentalized Diffusion Models (CDMs), where separate parameters (or adapters) are \textit{trained independently on different data sources, ensuring perfect (deterministic) isolation of their respective information}. All parameters are then merged at inference time and used jointly to generate samples. This technique is simple to implement with any existing DM architecture; CDMs are the first means to perform both selective forgetting (unlearning) and continual learning on large-scale diffusion models.

In addition to enabling the removal of information in the trained model from particular data, the method also allows attribution, which may inform the process of assessing the value of different cohorts of training data, as well as ensure that there is no memorization so the generated images are not substantially similar to those used for training.

The key enabler of CDMs is a closed-form expression for the backward diffusion flow as a mixture of the flows of its components, which is simple to derive and implement, but can suffer from two key problems. Implementation-wise, training and running inference with multiple models can quickly balloon the computational cost, and ensembling models trained on different subsets in principle can significantly underperform compared to a monolithic model, due to loss of synergistic information \cite{dukler2023safe}.

To address the first problem, we propose to use a pre-trained diffusion model and fine-tune on various downstream datasets. Fine-tuning helps the model preserve synergistic information across different shards \cite{dukler2023safe}. Further, to reduce the training/inference cost we can keep the single shared backbone fixed and train adapters \cite{hu2021lora} or prompt\cite{jia2022visual,sohn2023visual} on each disjoint shard of data.
Adapters can be trained remotely and shared with a central server without exposing the raw data, while prompts can use efficient batch-parallelization for quick inference. 

In regard to the latter problem, we empirically show that, in a variety of settings, a compartmentalized model can match the generative performance of a paragon model trained on all the data jointly (in some cases outperform a monolithic model), while allowing all the above mentioned data security improvements. This is both due to the particular objective of diffusion models, which in theory allows separate model training without any loss in performance (even if this need not be the case for real models), and to our use of a \textit{safe training set}, which allows the compartmentalized model components to still capture a significant amount of synergistic information \cite{dukler2023safe}.

\begin{figure*}[t]
    \centering
    \includegraphics[width=1.0\linewidth]{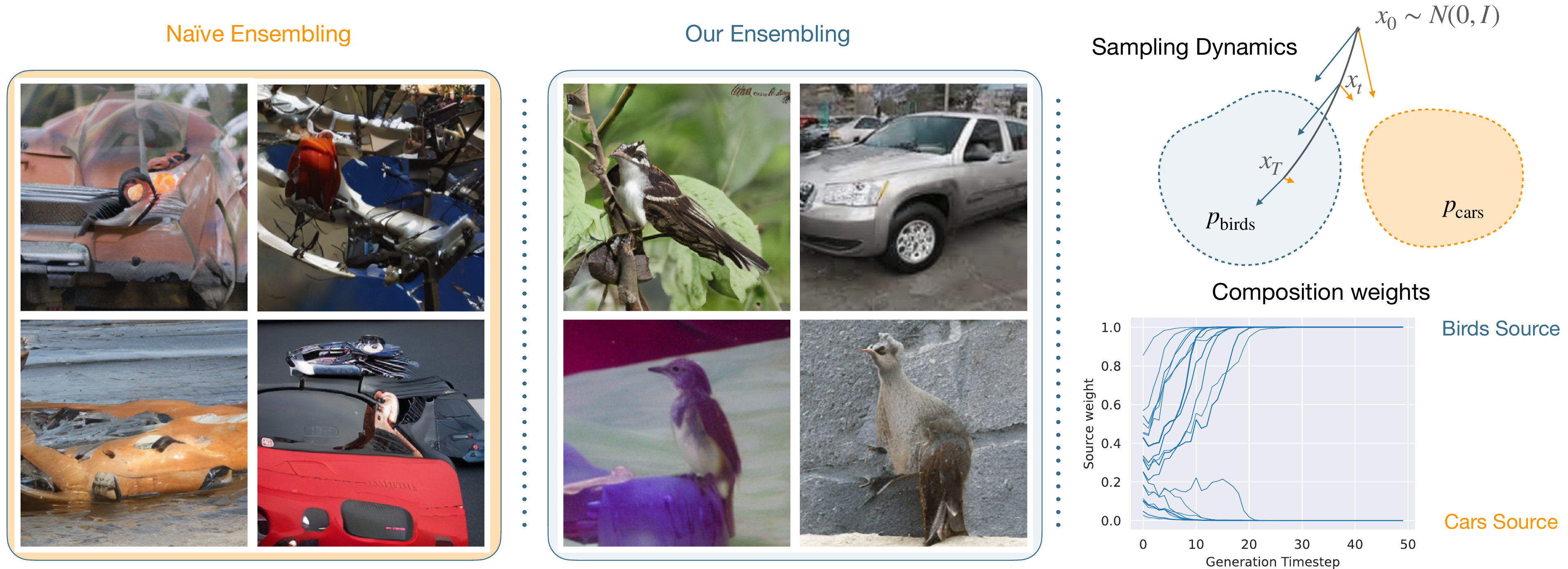}
    \caption{\textbf{Compositional diffusion models.} We train two diffusion models on two disjoint data distributions: Birds (CUB-200 \cite{wahcub}) and Stanford Cars \cite{krause20133d}. (Left) Image generated by naively composing the models by averaging their output. The sample images are distorted and contain elements of both distributions within the same image. (Center) Sample images generated by our method using the same models. The quality is substantially higher and the samples do not conflate the two distributions. (Right) The key idea is illustrated graphically at each step of  the reverse diffusion process, where we show the estimated optimal weights $w_i$ to assign to each component. At the beginning, the two components are weighted equally, also shown in the plot below, but as the model approaches convergence, the weights increasingly favor only one of the two models, based on the likelihood that it was trained on a data distribution closest to the current sample.}
    \label{fig:splash}
\end{figure*}

\section{Related Work}

\textbf{Forgetting/Unlearning: } Forgetting studies the problem of removing information pertaining to training data from the weights of a trained machine learning model. There are two major directions of works in forgetting, the first direction \cite{ginart2019making, bourtoule2021machine,yan2022arcane,kochno,kumar2022privacy,yu2022legonet, yan2022arcane,yu2022legonet,kochno, du2023reduce} involves splitting the training dataset into multiple shards and training separate models on each shard. This ensures that information contained in a particular training sample is restricted only to a specific subsets of parameters. When asked to remove a particular training sample, the unlearning procedure simply drops the corresponding shard and re-trains it without that sample. 

The second direction involves training a single machine learning model for the entire dataset, and providing approximate unlearning guarantees \cite{golatkar2020forgetting,Golatkar_2020_CVPR,mixedprivacyforgettinggolatkar}. Such methods rely on the linearization \cite{achille2021lqf} of the network with respect to a pre-trained initialization and then perform an approximate Newton step for stochastic forgetting \cite{golatkar2022mixed,guo2019certified}. \cite{neel2021descent,gupta2021adaptive,ullah2021machine,chourasia2022forget,sekhari2021remember, dwork2014algorithmic}.

\textbf{Diffusion Models: }Diffusion models are state-of-the-art generative models useful for high quality image generation \cite{ho2020denoising,song2020denoising,rombach2022high,dhariwal2021diffusion,lipman2022flow}.
\cite{rombach2022high,ramesh2022hierarchical} to video generation \cite{ho2022imagen,molad2023dreamix}. 
Diffusion models gradually add Gaussian noise to an image following a Markov process in the forward step during training to learn the score function, and perform denoising in the reverse step \cite{nelson1967dynamical,anderson1982reverse} to generate data using diffusion solvers \cite{lu2022dpm,lu2022dpm1,karras2022elucidating,song2020denoising}. \cite{song2020score} modelled diffusion models using stochastic differential equations (SDE). This enables the use of stochastic differential solvers and probability flow equations for reverse diffusion. 
\cite{bao2022all} uses a transformer based model using a ViT \cite{dosovitskiy2020image} which takes all in information (noisy image, timestep embedding, textual embedding) as input tokens different from standard diffusion models \cite{rombach2022high} which processes conditional information using cross-attention layers throughout the depth of the model. We use the U-ViT \cite{bao2022all} and Stable Diffusion \cite{rombach2022high} for experiments in this paper.

\textbf{Image Manipulation:}\cite{gandikota2023erasing,huang2023receler,kumari2023ablating,wu2024erasediff} proposed image manipulation techniques to prevent the diffusion model from generating certain concepts, however, such methods do not guarantee permanent removal of those concepts from the weights, which may be recovered through adversarial prompting. While CDMs certify removal of subsets of data/concepts as the corresponding sub-models are re-trained given an unlearning request. 

\textbf{Compositional Models, MoE:}\cite{du2023reduce,liu2022compositional,wang2023compositional} provided methods for compositional image generation, however their method is only aimed at improving the text-to-image alignment during generation. At inference, they propose to break the input prompt into subparts, compute the denoising prediction for each, and then average at each step during backward diffusion. While our method is aimed at improving the privacy of the model by sharding the training dataset into multiple subsets and training separate model for each. The two approach are completely orthogonal, as one involves breaking the inference prompt into nouns and using the same model multiple times, while ours involves splitting the training set and training separate models. Similarly mixture-of-experts (MoE) \cite{xue2023raphael,rajbhandari2022deepspeed} trains MoE layers with routing where each subset of parameters still contains information about the entire dataset (monolithic), and only a subset of parameters are used during inference to reduce computational cost. MoE lacks a a-priori separation of information in the weights making it unfavorable for perfect unlearning unlike CDMs.

\textbf{Memorization, Copyrights, and Differential Privacy:}The popularity of diffusion models has also prompted researchers to investigate memorization \cite{carlini2023extracting}, copyright protection \cite {vyas2023provable} and privacy in diffusion models. \cite{carlini2023extracting} showed successful extraction attacks on diffusion models raising privacy risks. \cite{vyas2023provable} provided a formalism for copyright protection in diffusion models using a less stringent version of differential privacy. They provided sampling algorithms to prevent the output of training samples from trained diffusion models after querying. To ensure privacy protected training of diffusion models, \cite{dockhorn2022differentially, ghalebikesabi2023differentially} proposed training diffusion models with differential privacy and show results toy datasets like MNIST/CIFAR \cite{lecun2010mnist, krizhevsky2009learning}.

In \Cref{sec:cdms} we propose compartmentalized diffusion models, shows its derivations, along with computation of the weights in \Cref{sec:weights}. Then we discuss the architecture and the implementation details in \Cref{sec:arch}, followed by the application of the proposed method in \Cref{sec:app} and conclude in \Cref{sec:conc}.

\begin{table*}[t]
    \centering
    \includegraphics[width=1\linewidth]{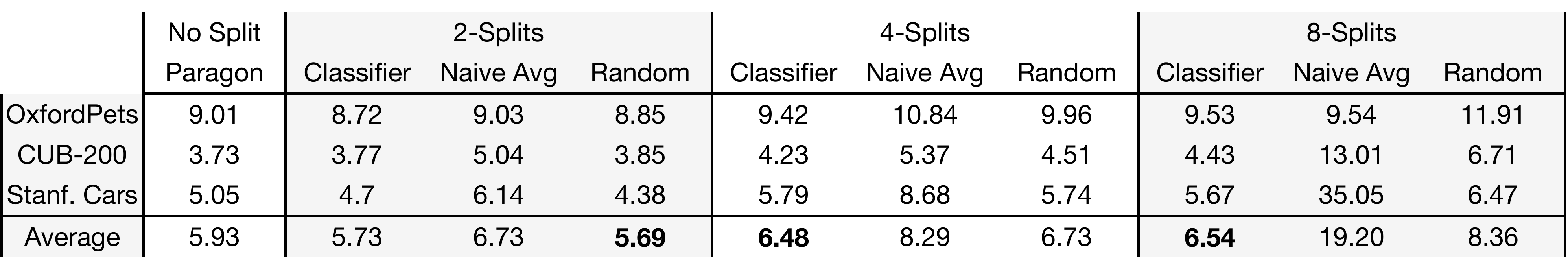}
    \caption{\textbf{Compartmentalized DMs for class conditional image generation.} We report, across various datasets, the FID score obtained with different methods to generate images starting from a compartmentalized model. We split each dataset uniformly across classes in multiple splits. Ideally the performance of the method should be close to the paragon performance of a non-compartmentalized model trained on all the data. We observe that for small number of shards the performance of the compartmentalized model can actually be better owning to the regularization effect of ensembling (FID score for 2-splits is lower (5.73, 5.69) compared to the paragon score 5.93). CDMs enable us split the training dataset into multiple shards with a minor increase in the average FID score (from 5.93 to 6.54 for 8-splits), while allowing us to unlearn much more efficiently.}
    \label{tab:class-cond}
\end{table*}

\vspace{-10pt}
\section{Compartmentalized Diffusion Models}
\label{sec:cdms}

Consider a dataset $\D = \{D_1, \ldots, D_n\}$ composed of $n$ of different data sources $D_n$. 
\textit{The core idea of CDMs is to train separate models or adapters independently on each $D_i$ to localize information, and compose them to obtain a model that behaves similarly to a model trained on the union $\bigcup \D_i$ of all data (monolithic paragon)}. We will use the score based stochastic differential equation formulation of diffusion models \cite{song2019generative}.

\vspace{-10pt}
\subsection{Diffusion models as SDEs}
Let $p(x_0)$ be the (unknown) ground-truth data distribution that we seek to model. At any time $t$ in the forward process, we define the conditional distribution of the input as $p_t(x_t|x_0) = \mathcal{N}(x_t;\gamma_tx_o, \sigma^2_tI)$, where $\gamma_t = \exp(-0.5 \cdot \int_0^{t}\beta_tdt)$ and $\sigma^2_t = 1 - \exp(-\int_0^{t}\beta_tdt)$. Using a variance preserving discrete Markov chain forward process, we obtain the following stochastic differential equation which models the forward process:
\begin{equation}
    dx_t = -\dfrac{1}{2} \beta_t  x_tdt + \sqrt{\beta_t}d\omega_t
\label{eq:forward-process}
\end{equation}
Here $x_t$ is the input at time $t$ in the forward process, $\beta_t$ are the transition kernel coefficients and $d\omega_t$ is the sandard Wiener process. Given the forward process, \cite{lindquist1979stochastic} showed that there exists a backward process, which enables us to generate samples from $p(x_0)$ given a random sample $x_T \sim \mathcal{N}(0,1)$. This is given by the backward diffusion equation

\begin{equation}
dx_t = \Big(- \dfrac{1}{2}\beta_t  x_t - \nabla_{x_t} \log p_t(x_t)\Big)dt + \sqrt{\beta_t}d\omega_t
\label{eq:reverse-process-sde}
\end{equation}

where $p_t(x_t) = \int_{x_0} p_t(x_t|x_0) p_0(x_0) dx_0$ is the marginal distribution at time $t$. \Cref{eq:reverse-process-sde} quite powerful as it highlights the fact that we only need access to $\nabla_{x_t} \log p_t(x_t)$ in order to generate samples from $p(x_0)$, which does not require access to the normalization constant. \cite{song2020score} also proposed an ordinary differential equation corresponding to \cref{eq:reverse-process-sde} which enables generating samples from $p(x_0)$ \cite{song2020denoising}. In practice, we model $p_t(x_t) = \int_{x_0} p_t(x_t|x_0) p_0(x_0) dx_0$ using a deep neural network $s_{\theta}(x_t, t)$ (or $\epsilon_{\theta}(x_t, t)$ as more commonly denoted in the literature\cite{ho2020denoising}), and optimize it using score matching \cite{song2019generative,song2020score,song2020sliced}.

\begin{figure*}[t]
    \centering
    \includegraphics[width=0.9\linewidth]{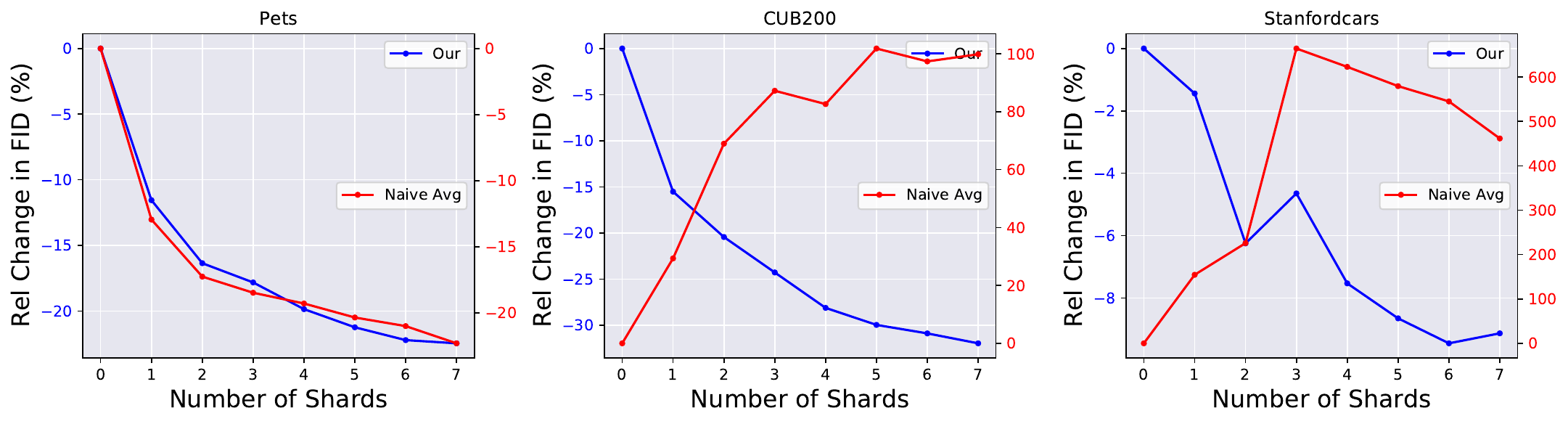}
    \caption{\textbf{Forgetting/Continual Learning with Compartmentalized DMs:} Relative change (wrt single shard) in the FID score as we continually add more shards (class-conditional generation with 8-splits). FID scores for our method continuously decreases as we add more shards compared to naive averaging which may result in incorrect mixture of vector fields. Classifier (our) based ensembling enables correct selection of models from the mixture of ensembles, and therefore results in decrease in FID with increase in data. This plot (when viewed from right to left) also shows the increase in FID, given a request to forget an entire shard.}
    \label{fig:forgetting}
\end{figure*}

\subsection{Compartmentalization}
Let us consider now the case where the data distribution $p(x_0)$ is composed of a mixture of  distributions:

\begin{equation}
p(x_0) = \lambda_1 p^{(1)}(x_0) + \ldots + \lambda_n p^{(n)}(x_0),
\label{eq:mixture-dist}
\end{equation}
Lets assume that we have access to the score functions for each of these $n$ distributions, $\{\nabla_{x_t}\log p^{(i)}(x_t)\}_{i=1}^n$, which can be used to generate samples. The key question is whether we can combine these mixture-specific score functions to generate a sample from the global distribution $p^{(i)}(x)$. To this end, we want to find the score function of the global distribution and write it using the score function of the individual distributions in \Cref{eq:global-marginal}. In practice, given the data sources corresponding to the mixture component $p^{(i)}(x)$, we sample $n$ training datasets $D_i$, and train $n$ independent diffusion model ($\{s^{(i)}_{\theta}(x_t, t)\}_{i=1}^n$) corresponding to each $p^{(i)}(x)$. Then using the trained models $s^{(i)}_{\theta}(x_t, t)$ we can approximate the empirical score for the global distribution and sample from it using diffusion samplers \cite{ho2020denoising,song2020denoising}.

To compute the score for the global distribution, we need to compute the global marginal distribution. Using the linearity of integration with a gaussian we can show that:
\begin{align}
p_t(x_t) &= \int p_t(x_t|x_0) \sum_{i=1}^n \lambda_i p^{(i)}(x_0) \nonumber \\
&= \sum_{i=1}^n \lambda_i p_t(x_t|x_0) p^{(i)}(x_0)
= \sum_{i=1}^n \lambda_i p^{(i)}_t(x_t)
\label{eq:global-marginal}
\end{align}

\subsection{Score of the mixture}
To sample from the global distribution \cref{eq:mixture-dist} using \cref{eq:reverse-process-sde} we need to compute the score of the marginal \cref{eq:global-marginal}. 

\begin{prop}
Let $\{s^{(i)}_{\theta}(x_t, t)\}$ be a set of diffusion models trained on $\{D_{i}\}_{i=1}^n$ separately. Then the score function corresponding to a diffusion model trained on $\{D_{i}\}_{i=1}^n$ jointly is given by,
\begin{equation}
s_{\theta}(x_t, t) = \sum_{i=1}^n w_t(x_t, t) s^{(i)}_{\theta}(x_t, t)
\label{eq:mixture-dm}
\end{equation}
where $w_t(x_t, t) = \lambda_i \dfrac{p^{(i)}_t(x_t)}{p_t(x_t)}$, $p_t(x_t) = \sum_{i=1}^n \lambda_i p^{(i)}_t(x_t)$.
We assume that each DNN has enough capacity, to minimize $\mathbb{E}_{x_0,t} \|\nabla_{x_t} \log p^{(i)}_t(x_t) - s^{(i)}_{\theta}(x_t, t)\|^2$. Thus we replace $\nabla_{x_t} \log p^{(i)}_t(x_t)$ with its empirical estimate $s^{(i)}_{\theta}(x_t, t)$. 
\label{prop:mixture-dm}
\end{prop}

\vspace{-20pt}
\subsection{Computing the weights}
\label{sec:weights}
The term $w_t(x_t, t)$ in \cref{eq:mixture-dm} has an intuitive interpretation. Let $x_0 \sim p(x) = \sum_i \lambda_i p^{(i)}(x)$ be a sample from the mixture distribution, and let $z \in \{1, \ldots, n\}$ be a discrete random variable which tells us the index of the mixture component that generated the sample (so that $p(x|z=i) = p^{(i)}(x)$ and $p(x) = \sum_i p(x|z=i)p(z=i)$. Then, by Bayes's rule, one readily sees that
\[
p_t(z=i|x) = \frac{p^{(i)}_t(x)}{p_t(x)}.
\]
That is, the additional weighting factor for each model can be interpreted as the probability that the current noisy sample $x_t$ originated from the data distribution used to train that model. To illustrate the behavior (see \Cref{fig:splash}), consider the case where $p^{(1)}(x)$ and $p^{(2)}(x)$ are disjoint (for example, images of pets and flowers respectively). At the beginning of the reverse diffusion, due to the amount of noise the sample is equally likely to be generated from either distribution, and both will have similar weight. As the time increases and more details are added to the sample, the image will increasingly be more likely to be either a pet or a flower. Correspondingly the generated image should draw only from the relevant domains, whereas using others would force the model to generate images of flowers by inductively combining images of pets (\Cref{fig:splash}).

This interpretation also gives us a way to compute $\frac{p^{(i)}_t(x)}{p_t(x)}$. In principle, one could estimate both $p^{(i)}_t(x)$ and $p_t(x)$ using the diffusion model itself, however this is computationally expensive. On the other hand, $p_t(z=i|x)$ is simple to estimate directly with a small auxiliary model. Let $f(x, t)$ be a $n$-way classifier that takes as input a noisy image $x$ and a time-step $t$ and outputs a $\mathrm{softmax}$. In this paper we try two classifiers, (1) k-NN using CLIP\cite{radford2021learning} for text-to-image models and (2) training a neural network classfier for class-conditional models. To train the network, we can generate pairs $\{(x_i, k_i)\}_{i=1}^N$ where $k_i \sim \operatorname{1, \ldots, n}$ is a random component index and $x_i \sim N(x| \gamma_t x_0, \sigma^2_t I)$, $x_0 \sim D_{k_i}$ is obtained by sampling a training image from the corresponding dataset $D_{k_i}$ and adding noise to it. The network is trained with the cross-entropy loss (standard image classification) to predict $k_i$ given $x_i$ and $t$. Then, at convergence $f(x, t) = \Big( \frac{p^{(1)}_t(x)}{p_t(x)}, \ldots, \frac{p^{(n)}_t(x)}{p_t(x)}\Big) = w_i(x_t, t)$, where $w_i(x_t, t)$ is from \cref{eq:mixture-dm}.

The classifier helps implement model selection at inference time, which aims to select the best model which describes the data distribution. However, when all the components of the mixture distribution are close in a distributional sense, we can replace the classifier, with \textit{naive averaging} of the ensemble of diffusion scores. In practice, using all the models at each time-step of backward diffusion can be computationally expensive, in such situations, we can approximate the averaging of scores, with simple \textit{random score selection}. Thus we have 3 methods for ensembling the diffusion scores at inference, (1) classifier, (2) naive averaging, and (3) random selection. We empirically show that classifier almost always outperforms naive averaging. Note that naive averaging may appear similar to \cite{du2023reduce, liu2022compositional, wang2023compositional}, however, there is one fundamental difference --  they use one fixed model and split the input prompt for composition, while naive averaging (CDMs in general) split the training data, train separate models (localize information) and perform compositional inference.

\begin{table*}[t]
    \centering
    \includegraphics[width=1\linewidth]{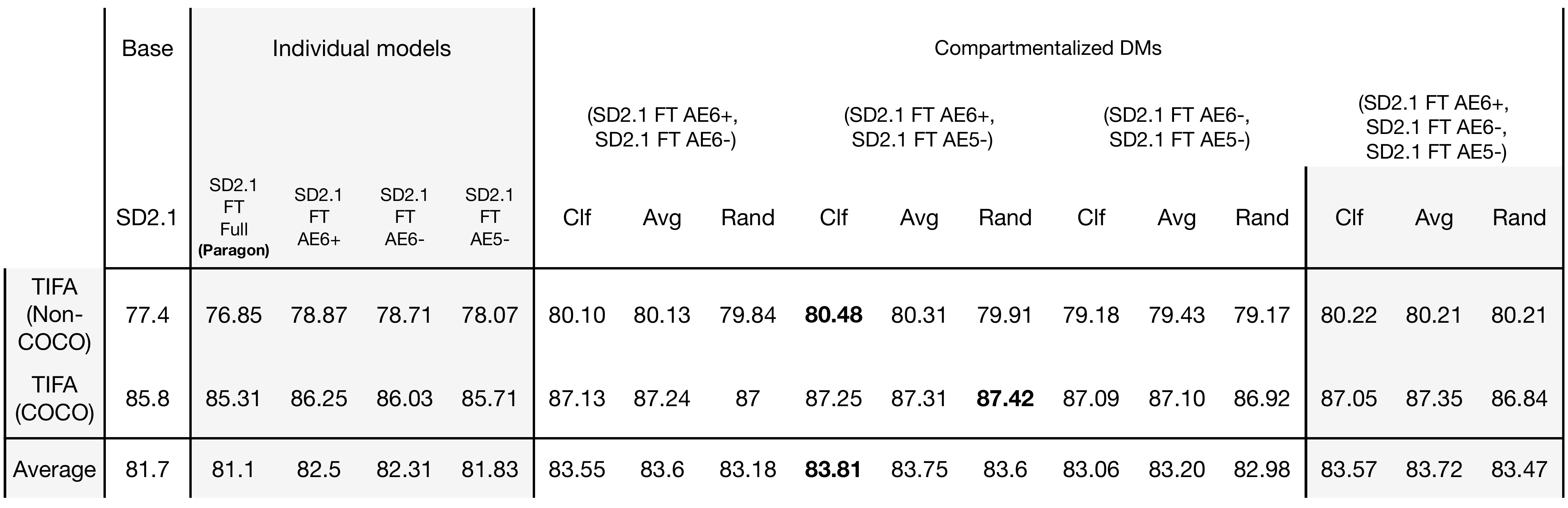}
    \caption{\textbf{Compartmentalized DMs improve text-to-image alignment}. We report the TIFA \cite{hu2023tifa}, text-to-image alignment score (higher is better) for different models. We obtain 3 subsets of MSCOCO \cite{lin2014microsoft} based on the aesthetic score inpsired from \cite{dai2023emu}. We obtain ~1k samples each with aesthetic score > 6.0 (AE6+), aesthetic score < 6 (AE6-), and aesthetic score < 5.0 (AE5-). We fine-tune SD2.1 on each of these subsets along with entire MSCOCO (full, paragon). We observe that fine-tuning SD2.1 (individual models) helps improve alignment compared to the base model. However, CDMs further improve the alignment, from 81.7 for the base model to 83.81 for the best CDM. Even the worst CDM has better performance compared to the best individual model, and paragon. This shows the regularization effect provided by CDMs, compared to fine-tuning a single model on the entire dataset.}
    \label{tab:alignment}
\end{table*}

\section{Architecture and Implementation}
\label{sec:arch}
We use Stable Diffusion 2.1 Base (SD2.1) \cite{rombach2022high} for text-to-image generation (512 $\times$ 512), and U-ViT \cite{bao2022all} for unconditional and class conditional generation ($256 \times 256$). CDMs can be used to compose any set of diffusion models, so long as the output space is same for all the models. If the number of shards increase arbitrarily, then fine-tuning adapters (like prompts) improves storage and inference efficiency. We perform both set of experiments, for text-to-image generation we fine-tune the entire model (SD2.1) while for class-conditional generation we fine-tune adapters (U-ViT).

\textbf{Datasets} We use MSCOCO \cite{lin2014microsoft} as the private dataset for training text-to-image CDMs, and fine-grained datasets like CUB200 \cite{wahcub}, Stanfordcars \cite{krause20133d}, OxfordPets \cite{parkhi2012cats} for class-conditional models.  We split MSCOCO based on the aesthetic score of the images, and the fine-grained datasets based on the class label of the images where each split can be considered as data from a separate user.

\textbf{Classifier}
For text-to-image generation we use a k-NN classifier with CLIP embeddings. More precisely, at each $t$ in backward diffusion, we predict $x_0$ using the diffusion model, which is used by the k-NN classifier, to compute probability scores for each data source. For unconditional and class-conditional image generation, we train an neural network (linear layer plus attention block) on top of intermediate level features of a U-ViT to predict the classifier weights for different data sources. Since Stable Diffusion is pre-trained on a much larger data source compared to the U-ViT, we observe that is sufficient to use a k-NN classifier instead of training a new neural network.

\section{Applications}
\label{sec:app}
CDMs empower users to selectively incorporate or exclude subsets of training data, achieving performance levels comparable to training monolithic models on the complete dataset. Additionally, these models impart a regularization effect, enhancing the alignment between the textual and visual elements, all the while facilitating subset attribution. When all the weights in CDMs are uniformly assigned (naive averaging), it mitigates memorization, thereby satisfying the guarantee of copyright protection \cite{vyas2023provable}. We will elaborate on each of the applications of CDMs in this section. 

\begin{figure*}[t]
    \centering
    \includegraphics[width=0.99\linewidth]{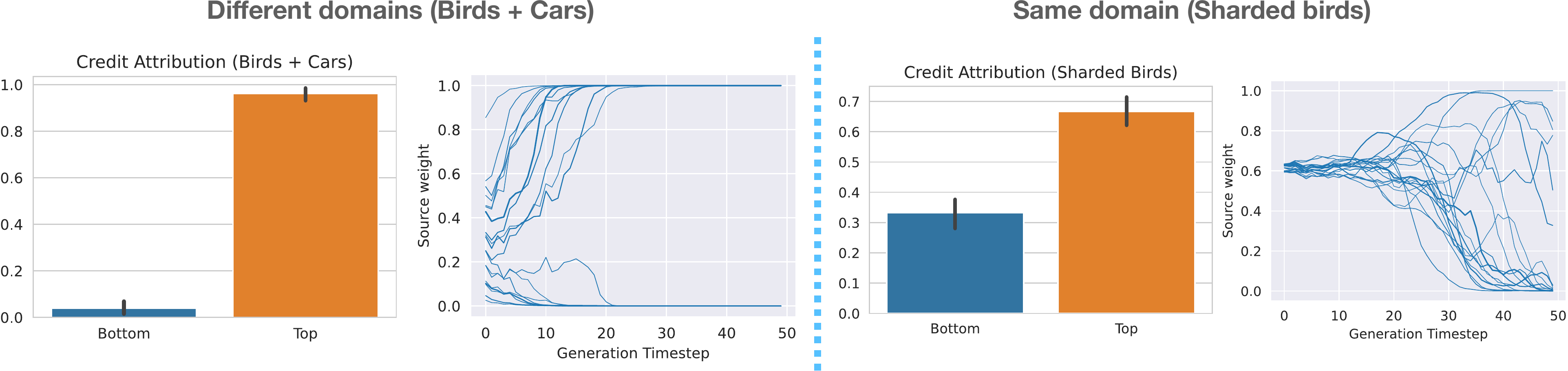}
    \caption{\textbf{Credit attribution with CDMs} Compartmentalized DMs enable us to provide credit attribution corresponding to samples belonging to different models. Plot shows average credit attribution when shards are from different domains (CUB200 \cite{wahcub} and Cars \cite{krause20133d}) and when shards are uniform split of the same domain. For different source domains, CDMs selects the appropriate domain model during backward diffusion, resulting in higher source weight for one model compared to another. For same source domains, CDMS assigns equal weight for majority of backward process, until the end when it selects one source model.}
    \label{fig:credit}
\end{figure*}

\begin{figure*}[t]
    \centering
    \includegraphics[width=0.99\linewidth]{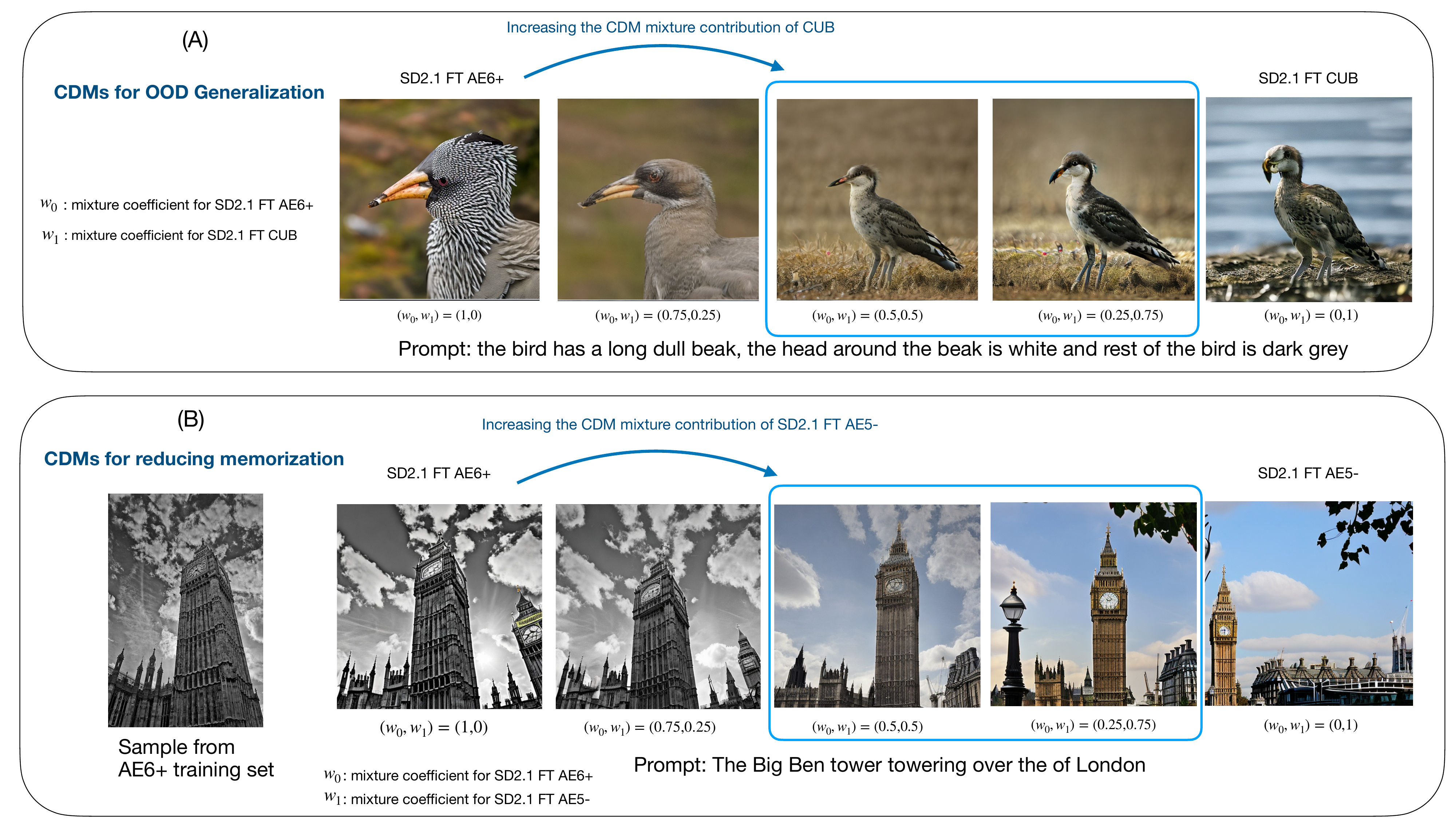}
    \caption{\textbf{Out-of-Distribution Coverage and Memorization:} Top (A): CDMs enable out of distribution (OOD) generalization by combining shard belonging to different domains. Figure (A) shows that SD2.1 fine-tuned (FT) on AE6+ produces unrealistic images of the bird, and does not follow the prompt correctly, however adding a shard corresponding to birds (SD2.1 FT CUB), with appropriate mixture weights enables the CDMs to generalize OOD, and produce more realistic birds (eg $(w_0,w_1)=(0.5,0.5)$ or $(w_0,w_1)=(0.25,0.75)$. CDMs have the flexibility to add diffusion score from different data sources to improve OOD generalization. Bottom (B): CDMs can also help in reducing memorization. Figure (B) shows that SD2.1 fine-tuned (FT) on AE6+ often memorizes training data \cite{carlini2023extracting}, eg Big Ben tower $(w_0,w_1)=(1,0)$, which can often violate the privacy rights of the user data. We can prevent memorization in diffusion models by using a mixture of models like in CDMs, which will merge diffusion flows from multiple sources preventing generation of memorized training samples at inference (eg $(w_0,w_1)=(0.5,0.5)$ or $(w_0,w_1)=(0.25,0.75)$). We show that CDMs also aid improving the diversity of the generated images, while preserving the inherent concept presented in the input prompt.}
    \label{fig:ood-mem}
\end{figure*}

\textbf{Forgetting.} Owners of the training data may, at any point, modify their sharing preferences leading to a shrinking set $S$ of usable sources. When this happens, all information about that data needs to be removed from the model. However, the large amount of current state-of-the-art diffusion models precludes re-training on the remaining data as a viable strategy. Compartmentalized models such as CDMs allow for a simple solution to the problem: if a data source $D_i$ is removed, we only need to remove the corresponding model to remove all information about it. Moreover, if only a subset of a training source is removed, it is only necessary to retrain the corresponding model. We show that increasing the number of splits does not increase the FID scores after composition (\Cref{tab:class-cond}) which is critical for forgetting as its enables easy removals of shards without significantly loosing performance. \Cref{fig:forgetting} shows the relative change in the FID score as we drop shards.

\textbf{Continual Learning.} The data sources $D_i$ may represent additional batches of training data that are acquired incrementally. Retraining the model from scratch every time new data is acquired, or fine-tuning an existing model, which brings the risk of catastrophic forgetting, is not desirable in this case. With CDMs, one can simply train an additional model on $D_i$ and compose it with the previous models. In \Cref{fig:forgetting} we show that adding more shards in a continual fashion improves the FID score relative to a single shard. Also, simple naive averaging over the shards will results in incorrect mixture of vector fields which can be avoided by the method proposed in \Cref{prop:mixture-dm}.

\textbf{Text-to-Image Alignment} \cite{dai2023emu} showed that fine-tuning diffusion models on high quality data improves text-to-image alignment. In \cref{tab:alignment} we show that fine-tuning diffusion models in a compartmentalized fashion provides much better alignment (83.81 TIFA score) compared to fine-tuning a single model on the joint data source (81.1 TIFA score). We obtain 3 subsets of MSCOCO based on aesthetic scores, (1) AE6+: ~1k samples with aesthetic score > 6, (2) AE6-: top ~1k samples with aesthetic score < 6, (3) AE5-: top ~1k samples with aesthetic score <5. We fine-tune SD2.1 on each data source, and compose them with CDMs. We observe that CDMs consistently outperform the individual models, paragon, and base SD2.1 model for all composition methods. This shows CDMs not only enable easy unlearning of any subset of MSCOCO, but also improve alignment due to the regularization effect of ensembling.

\textbf{Measuring contribution of individual sources.} Let $x_0$ be a sample generated solving the ODE \cref{eq:reverse-process-sde} starting from an initial $x_1 \sim p_1(x)$. The likelihood of a generated image can then be computed as
\[
\log p_1(x_1) - \log p(x_0) =  - \int_0^1 \div \nabla_{x_t} \log{p^{(i)}(x_t)} dt,
\]
that is, the divergence of the score function integrated along the path. In the case of a CDM, this likelihood can further be decomposed as:
\begin{align}
&\log p_1(x_1) - \log p_0(x_0)  = \sum_i \lambda_i L_i \nonumber \\
&= \sum_i   \lambda_i \int   \div \big( w_i(x_t, t) \, \nabla_{x_t} \log{p^{(i)}(x_t)} \big) dt
\label{eq:attribution}
\end{align}
where $L_i$ can be interpreted as the contribution to each component of the model to the total likelihood. Using this, we can quantify the credit $C_i$ of the data source $D_i$ as:
\[
C_i = \frac{\lambda L_i}{\sum_{j=1}^n \lambda_j L_j}.
\]

We note that while $\sum_i \lambda_i L_i$ is the likelihood assigned by the CDM to the the generated sample, one cannot interpret the individual $L_i$ as the likelihood assigned by each submodel. In \Cref{fig:credit} we show that when shards belongs to different distributions the credit attribution is correctly more skewed (generated image belongs to one distribution) compared to similar distributions which has a more uniform attribution (since all distributions are similar). The composition weights for different domains at inference start with similar values and change rapidly within the first 10 generation steps (see \Cref{fig:credit} left). For same domains the weights start with similar values and maintain them until almost half generation is complete before selecting one split (\Cref{fig:credit} right).

\textbf{Better out-of-domain (OOD) coverage and reduce memorization} Often times diffusion models under-perform on certain sub-populations of the training data. For eg. in \cref{fig:ood-mem} (A) we show that SD2.1 fine-tuned on MSCOCO AE6+ (with TIFA alignment score of 82.5, see \cref{tab:alignment}) is unable to produce realistic birds (OOD sub-population) when provided with descriptive prompts. However, we show that using CDMs we can compose SD2.1 FT AE6+ (in \cref{fig:ood-mem}) with SD2.1 FT CUB-200 (birds dataset) at inference to obtain improved alignment, better OOD coverage, with realistic looking birds. In \cref{fig:ood-mem}, $w_0, w_1$ correspond to mixture weights from \cref{prop:mixture-dm}.

In \cref{fig:ood-mem}, we show that diffusion models tend to memorize training data \cite{carlini2023extracting}. CDMs can reduce memorization in diffusion models by ensembling diffusion paths from different models at inference, as a result the generated image will not resemble output from any particular source model. CDMs help improve the diversity of the synthesized images along with reduced memorization. This is because using naive averaging is equivalent to sampling from Algorithm 3 in \cite{vyas2023provable} which provide copy protection, and thus reduces memorization.

\textbf{Limitations}
Even though CDMs enjoy a myriad of nice properties like easy unlearning, continual model update, credit attribution, improved alignment, OOD coverage and reduce coverage, they suffer from increase in number of training parameters, and high inference cost. Increased parameters, and inference compute can be reduced by the use adapters at the expense of model performance, however, it cannot be completely eliminated. Random selection of scores in CDMs provide an efficient way to reduce the compute requirements. Application of CDMs is simplified in situations when the data is naturally partitioned by the user privacy rights, however, in other situations sharding the data in a manner which preserves the synergistic information after compartmentalization is challenging (one can always split uniformly).

\section{Conclusion}
\label{sec:conc}
Data protection is an increasingly arduous task as the volume of training data needed to train massive AI models increases. While techniques to manage privacy and attribution have been demonstrated for a variety of model architectures, mostly at relatively small scale, up to now it was not possible to directly apply them to Diffusion Models. We present the first method to compose such models in a private manner, and illustrate its use in
selective forgetting, continual learning, out of distribution coverage, reducing memorization, credit attribution, and improving alignment. We show that we can train compartmentalized diffusion models for deep networks (or adapters) to model different data distributions, and perform comparable (or even better) to a model trained on the joint distribution. CDMs also provide a natural way for
customized model inference (`a-la-carte) \cite{bowman2023carte} which enables user to arbitrarily choose a subset of shards at inference time, provides a way for copyright protected generation \cite{vyas2023provable}, and encourage exploring differentially private adapter tuning for diffusion models. Increasing the number of shards for CDMs in the limit will lead to retrieval augmented diffusion models, which further helps in privacy protected generation as samples can be easily removed (unlearning) or added on the fly, while providing credit attribution, and more synthesis control with the retrieved samples. 

\section{Impact Statements}
This paper presents work whose goal is to provide a novel method for training diffusion models through compartmentalization with several privacy benefits. If applied at scale, this has a lot of societal consequences, for instance, it will allow diffusion model user to make unlearning request without requiring to discard the entire model, provide subset attribution to users, and prevent sampling of memorized training data, thus promoting safe use of diffusion models and build user trust in AI.

{
    \small
    \bibliographystyle{ieeenat_fullname}
    \bibliography{main}
}

\clearpage
\appendix
\onecolumn

\begin{center}
\begin{Large}
Appendix
\end{Large}
\end{center}

\section{Proof of \cref{prop:mixture-dm}}
\begin{proof}
\begin{align}
    \nabla_{x_t} \log p_t(x_t) &= \nabla_{x_t} \log \Big({\sum_{i=1}^n \lambda_i p^{(i)}_t(x_t)}\Big) \nonumber \\
    &= \dfrac{1}{\sum_{i=1}^n \lambda_i p^{(i)}_t(x_t)} \sum_{i=1}^n \lambda_i \nabla_{x_t} p^{(i)}_t(x_t) \nonumber \\
    &= \dfrac{1}{p_t(x_t)} \sum_{i=1}^n \lambda_i \nabla_{x_t} p^{(i)}_t(x_t) \dfrac{p^{(i)}_t(x_t)}{p^{(i)}_t(x_t)} \nonumber \\
    &=  \sum_{i=1}^n \lambda_i \dfrac{p^{(i)}_t(x_t)}{p_t(x_t)} \nabla_{x_t} \log p^{(i)}_t(x_t)  
\end{align}
\end{proof}

\section{Additional Discussion}
\textbf{How is CDMs method different from the existing works on compositional diffusion models like \cite{du2023reduce,liu2022compositional,wang2023compositional}?} Existing works on compositional generation with diffusion models are aimed at improving the text to image alignment of the models, and attempt to improve the composition of different objects in the scene. This is achieved by either manipulating the cross-attention layers \cite{wang2023compositional} of the model architecture to re-weight the attention matrix weights to increase the contribution of objects or nouns in the input prompt, or by sampling flows for each noun (or keywords) separately and combining them for better composition \cite{du2023reduce,liu2022compositional}. On the other hand CDMs compartmentalize the training data to improve the privacy guarantees of the model. This involves controlled isolation of information about the data into different parameters (adapters), and composing them arbitrarily at inference time based on the users access rights. One major benefit of this approach is the ability to unlearn subsets of data efficiently compared to a monolithic model. Given a forgetting request by a user, CDMs can unlearn a subset by simply re-training the shard (from the last checkpoint before those samples were seen) to ensure perfect unlearning.

\textbf{Is CDMs the same as MoE\cite{xue2023raphael,rajbhandari2022deepspeed}?} MoE aims at increasing the model size (hence performance) while keeping the inference cost constant (or low). This involves training MoE layers with dynamic routing to select a subset of parameters at inference based on the input. At the core of MoE lies the MoE layer which has a set of shared parameters, and a set of disjoint expert parameters. The shared parameters are updated with gradients from the complete dataset, while the experts are updated based on subsets of data based on the routing scheme (which can be data dependent). This leads to mixing of information between different subsets of data into the parameters of the model making it equivalent to a monolithic model from an unlearning standpoint. Furthermore, if the routing scheme is data-dependent i.e. it was learned using the complete data, then each unlearning request will require the re-training the routing scheme. CDMs simplify this problem, by training different parameters for different subsets of data, and moreover provide a closed form solution for the output score function as a combination of the individual scores. 

\textbf{Is CDMs different from a monolithic model if the unlearning request necessitates re-training of all the shards?} One of the important problems in machine unlearning is -- how to shard the training data -- whether it should be split uniformly at random or based on certain properties or statistic. Sharding plays critical role in determining the computational cost of unlearning given a forgetting request. \cite{koch2023no} proposes to identify subsets of data which high likelihood of forgetting, and keep them in a separate shard or use it to update the weights toward the end of training. \cite{dukler2023safe} further builds on it a proposed to construct a shard graph where each node is small subset of data, and the edges correspond to different data owner rights. For instance, data should be clustered according to the source, such that once a source decides to withdraw the remaining shards are unaffected. Similarly different sources may have different access rights with respect to other sources which can be captured in a graph, and the cliques can be used to train different models. In our experiments we attempt to mimic this by sharding the fine-grained data based on its class label where each class is a different source, and MSCOCO based on its quality as we may want to localize images based on the quality to identify its value.  The aim of an ideal system enabling unlearning should be to device a sharding mechanism which will reduce the number of shards to be re-trained given an unlearning request on an average given a series of requests. In the worst case setting re-training of all the shards may be warranted to ensure unlearning (in which case its equal to monolithic), however, in the average case compartmentalization triumps a monolithic model (for eg. $8\times$ computational improvement for CDMs in our experiments).

\textbf{What is the relation between CDMs and copy protection\cite{vyas2023provable}?} Copy protection has recently received increased attraction with the burst of generative models. \cite{vyas2023provable} proposed a definition of copy protection by measure the different of a generative model to a safe model which was not trained on protected data. They also propose the CP$-\Delta$ algorithm (Algorithm-3) for copy protected generation which involves splitting the training data into two shards, training separate models on each and sampling from the product of the distribution. While sampling from the product of distribution is easy for language models, its non-trivial for diffusion models. However, we can circumvent this issue for diffusion using CDMs by simply computing the score of the copy protected distribution from Algorithm-3 \cite{vyas2023provable}. Given two models $p_1$ and $p_2$ trained on disjoint subsets of data, \cite{vyas2023provable} proposes to sample from $\sqrt{p_1(x)p_2(x)}/Z$. Computing the score of distribution we obtain $\dfrac{1}{2}(\nabla_x \log p_1(x) + \nabla_x \log p_2(x))$, where $\nabla_x \log p_1(x), \nabla_x \log p_2(x)$ is often approximated in practice by trained diffusion models. Looking at this equation carefully -- we realize that it is equivalent to CDMs when we use naive-averaging instead of the classifier (i.e. equal weights for all the model). This is depicted pictorially in \cref{fig:ood-mem}. 

\textbf{Can CDMs provide/quantify sample attribution?} While providing attribution to a single sample in the training set during inference in extremely difficult (due to the non-convexity and non-linearity of diffusion models) CDMs can provide and quantify sample attribution using \cref{eq:attribution}. Source attribution improves the explainability and transparency of diffusion models as they provide users with the ability to localize the contribution training samples to a source, which can further be used to reduce bias and improve fairness of the model. In practice, the source attribution can be efficiently computed using the method proposed in \cite{kong2023information}. 

\section{Implementation Details}
We use latent diffusion models for all the experiments in the paper. Latent diffusion models decrease the computational complexity of the model while preventing decrease in quality. We use Stable diffusion diffusion models for experiments on text-to-image generation and transformer based diffusion models for all other experiments performed in the paper. We use the Huggingface based implementation of Stable diffusion and for transformer based experiments we use the U-ViT architecture proposed in \cite{bao2022all}, whose code can be publicly found here: https://github.com/baofff/U-ViT. 

To reduce the computational complexity of the compartmentalized models at inference,  we train adapters (prompts or LoRA) to learn the data distribution. For transformer based experiments we use an ImageNet pretrained class-conditional model as the backbone and learn prompts (adapters) to model the data distribution. We use the Adam optimizer with a learning rate of 0.1 for transformer based experiments and 1e-4 for Stable diffusion based experiments. We do not use weight decay while fine-tuning and use a batch-size of 256 for all experiments. 

For unconditional image generation we train one set of deep prompts (deep prompts: prompts appended to all the layers of a the network, see \cite{jia2022visual}) for the entire distribution. We use a prompt length of 256 tokens, and train for 800 epochs. For class-conditional image generation, we train one set of deep prompts for each class. Since we train class-wise prompts, we use a prompt length of 8 tokens, and again train for 800 epochs. For text-to-image generation, we fine-tune the model for around 15000 steps.

For training the classifier, we use k-NN based classifier for Stable diffusion model, and model based classifier for transformer based experiments. For k-NN based classifier we use CLIP features to generate k-mean embeddings for each shard of the data. During inference, at each time-step, we compute $x_0$ from each $x_t$, and then compute the normalized (soft-max) distance to each shard.

\section{Additional Figures}
\begin{figure}[h]
    \centering
    \includegraphics[width=0.99\linewidth]{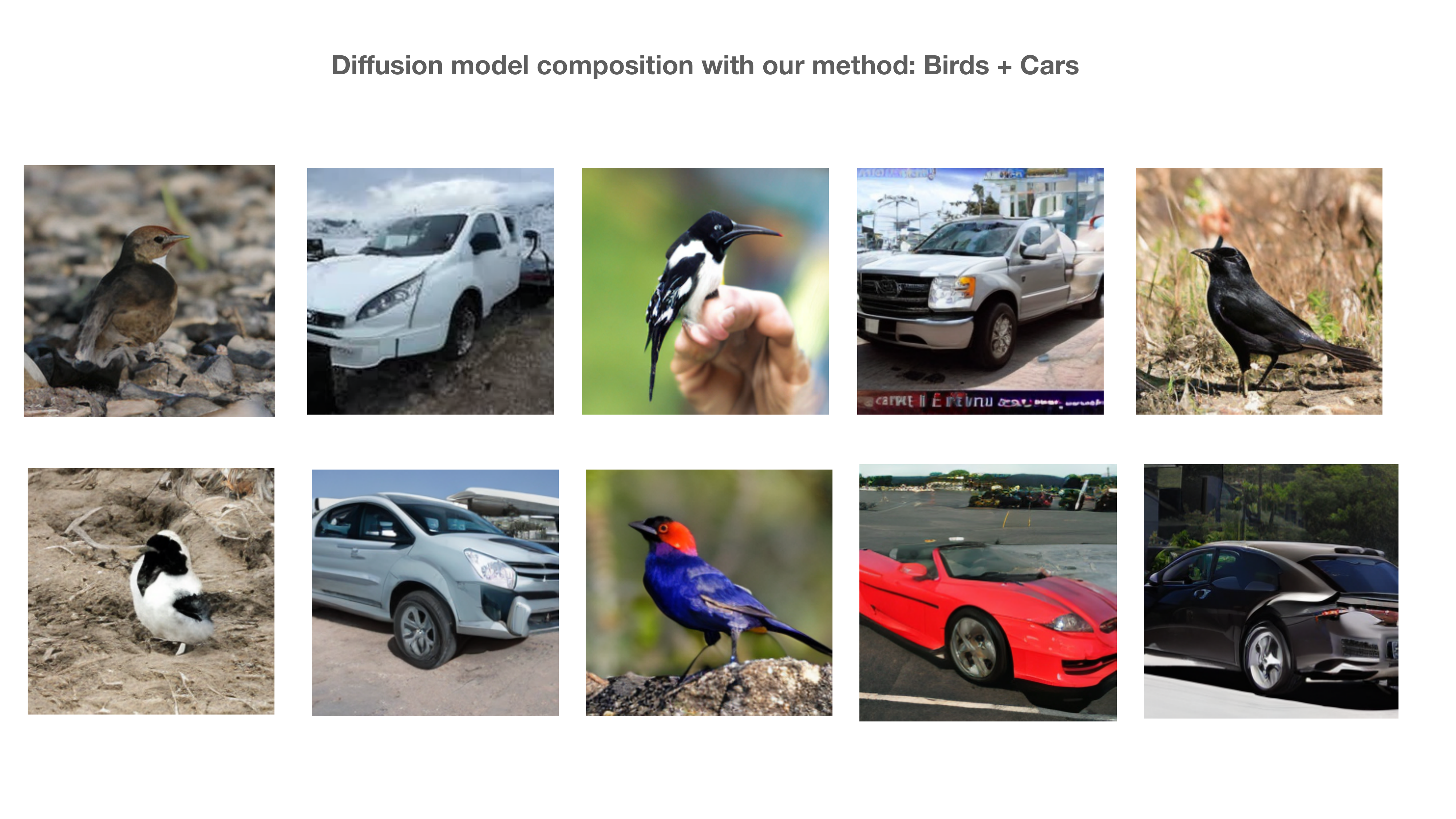}
    \includegraphics[width=0.99\linewidth]{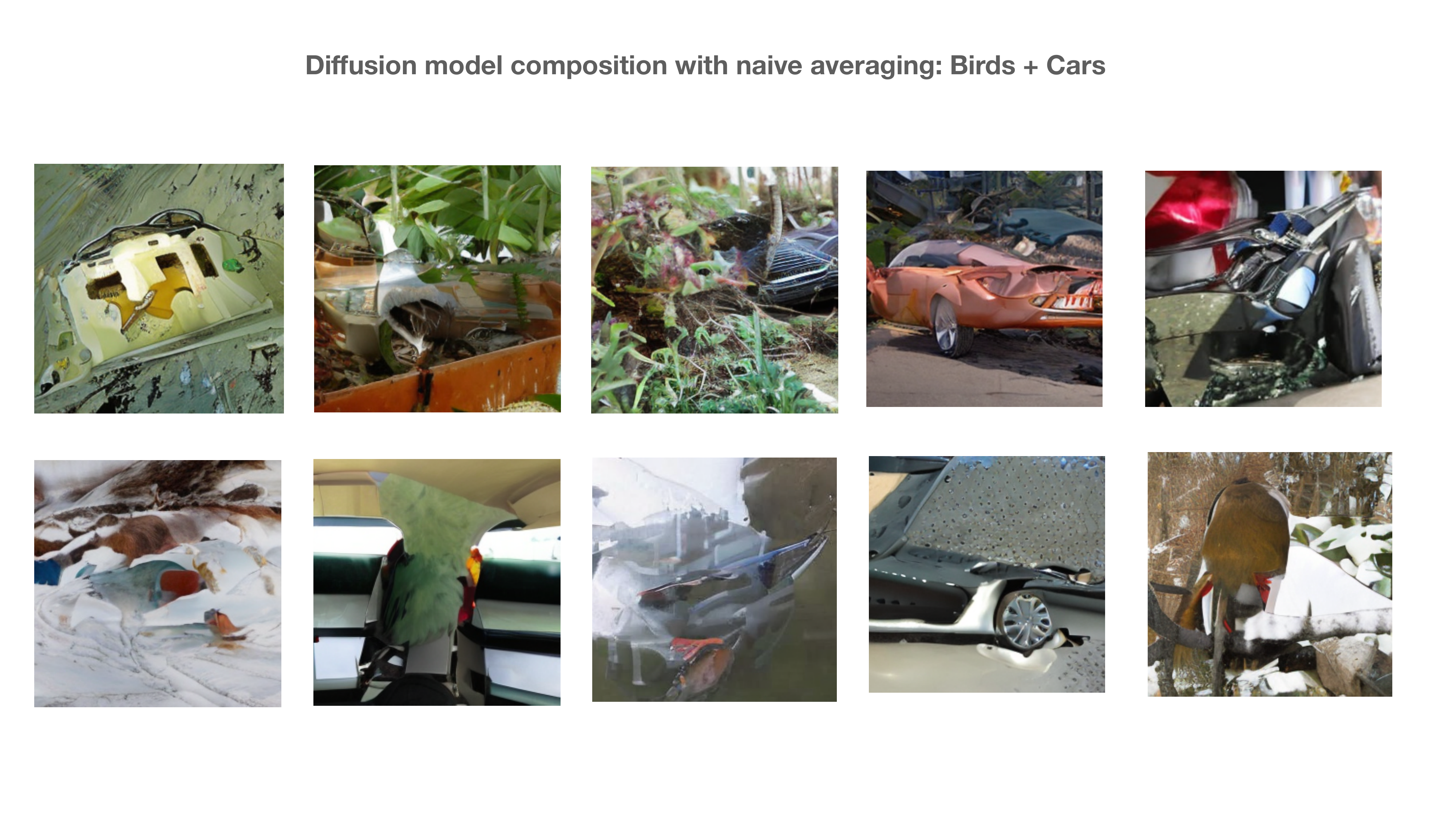}
    \caption{Diffusion model composition with our method vs naive averaging}
\end{figure}

\begin{figure}[h]
    \centering
    \includegraphics[width=0.99\linewidth]{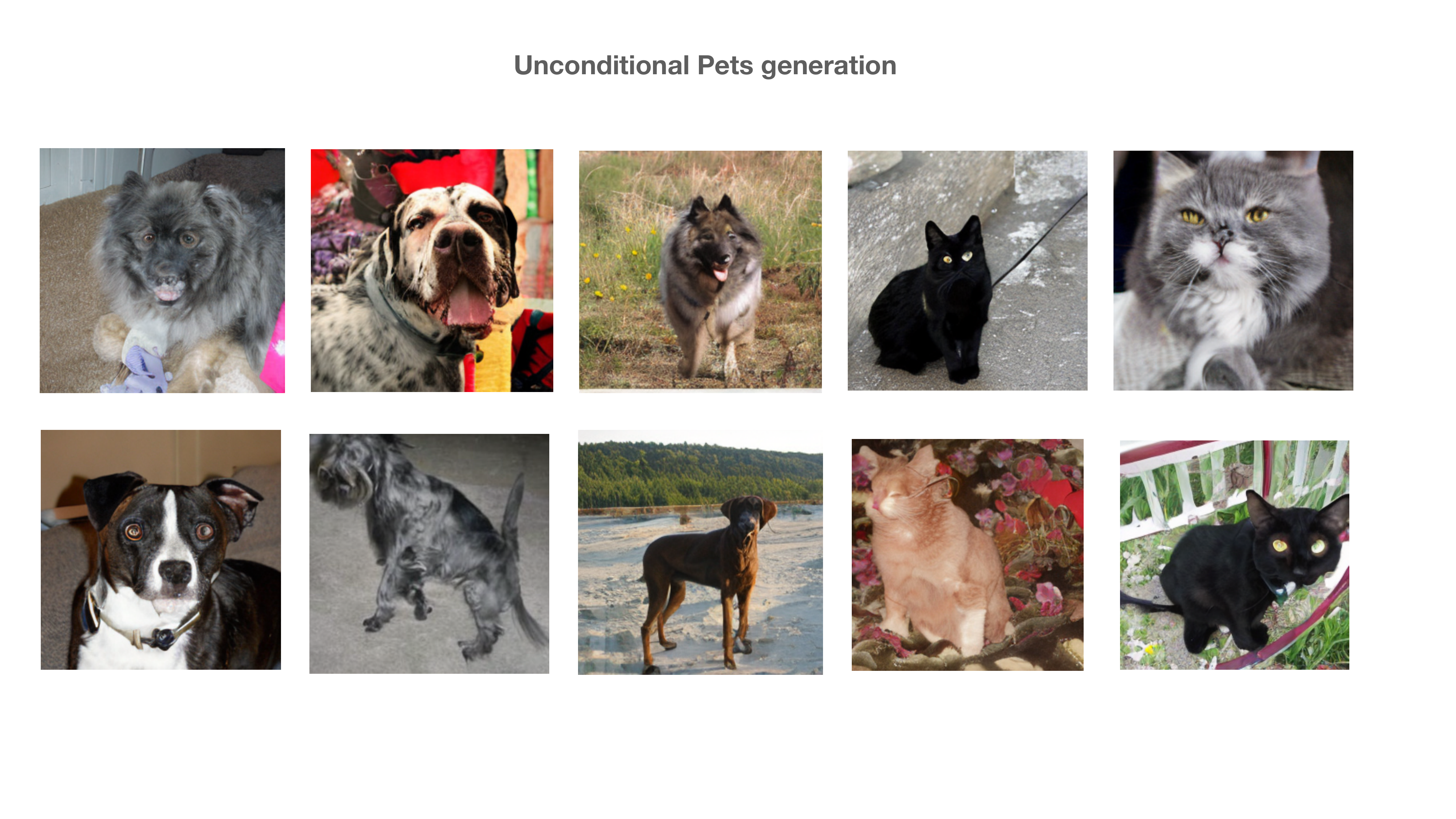}
    \caption{Unconditional image generation with a prompt (adapter) based model trained on OxfordPets}
\end{figure}

\begin{figure}[h]
    \centering
    \includegraphics[width=0.99\linewidth]{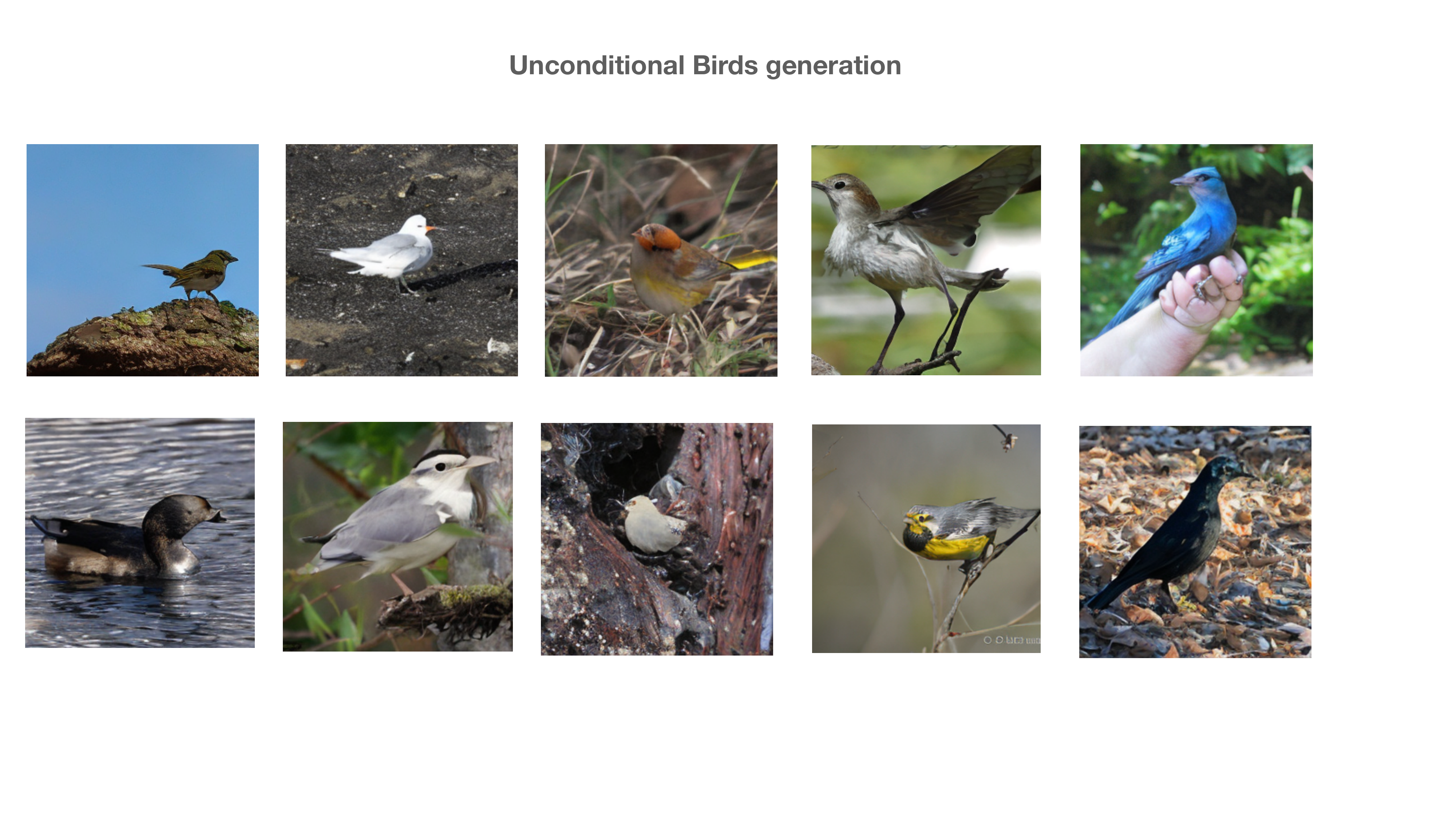}
    \caption{Unconditional image generation with a prompt (adapter) based model trained on CUB200}
\end{figure}

\end{document}